\documentclass[letterpaper, 10 pt, conference]{ieeeconf}  
\IEEEoverridecommandlockouts                              %
\usepackage{graphics} 
\usepackage{epsfig} 
\usepackage{times} 
\usepackage{amsmath}
\usepackage{amssymb}
\usepackage{cite}
\usepackage{float}
\usepackage{booktabs}
\usepackage{siunitx}
\usepackage{threeparttable}
\usepackage{subcaption}

\usepackage{xspace}
\usepackage{xcolor}
\newcommand{\cpnav}{\textsc{CPNav}\xspace}

\title{\LARGE \bf
\cpnav: Calibrated Vision–Language Navigation for Object-Goal Search in Unseen Environments
}

\author{%
Qizhao Chen$^{1}$, Yuanhong Zeng$^{2}$, Shoh Nishino$^{1}$, and Anushri Dixit$^{1}$%
\thanks{$^{1}$Qizhao Chen, Shoh Nishino, and Anushri Dixit are with the Department of Mechanical and Aerospace Engineering, University of California, Los Angeles, Los Angeles, CA, USA.
{\tt\footnotesize qizhaoc@ucla.edu, shohnishino@ucla.edu, anushridixit@ucla.edu}}%
\thanks{$^{2}$Yuanhong Zeng is with the Department of Electrical and Computer Engineering, University of California, Los Angeles, Los Angeles, CA, USA.
{\tt\footnotesize yuanhongzeng@ucla.edu}}%
}

\usepackage{hyperref}
\hypersetup{pdfauthor={},pdftitle={}}

\begin{document}

\maketitle
\thispagestyle{empty}
\pagestyle{empty}

\begin{abstract}

Directing a robot to locate a target object in unseen environments remains a central challenge in embodied AI. Vision--Language Model (VLM)-based methods provide rich semantic reasoning but often produce uncalibrated and overconfident action decisions. We introduce \cpnav, a navigation framework that combines collision-aware planning with statistically calibrated VLM action scoring for map-free object-goal navigation. From RGB-D input, \cpnav constructs a navigability map, generates collision-free motion primitives, and queries two VLMs for goal detection and action scoring. The resulting scores are calibrated offline via conformal prediction (CP) using oracle actions obtained through a tree-structured exploration procedure with backtracking, providing finite-sample guarantees at a user-specified risk level. The CP layer is model-agnostic and can wrap any VLM policy. Online, calibrated VLM scores guide action selection, while a tree-structured exploration history records alternative actions and allows backtracking. We evaluate \cpnav in both simulation and real-world hardware settings. In AI Habitat simulation, compared with the uncalibrated baseline method, \cpnav achieves relative SR and SPL improvements of 6.2\% and 11.4\% on HM3D v0.2, and 12.4\% and 8.8\% on MP3D, respectively. In real-world deployment on a modified Hiwonder MentorPi robot, the corresponding improvements in SR and SPL are 24.9\% and 15.2\%, respectively. Code is available at \href{https://github.com/practice-lab-ucla/CPnav}{\texttt{github.com/practice-lab-ucla/CPnav}}.
\end{abstract}

\section{Introduction}
Robotic navigation in complex unseen indoor environments remains challenging because effective decision-making must tightly couple low-level path planning with high-level semantic understanding, while generalizing to novel layouts without prior maps or goal location information \cite{SemanticNavSurvey}. Traditional rule-based planners can handle geometric obstacle avoidance but rely on hand-tuned rules and expert knowledge, making them brittle in unforeseen situations \cite{cadena2016past}. In contrast, learning-based navigation planners, particularly those based on reinforcement learning (RL), often require painstaking reward shaping and large-scale training data \cite{dulac2019challenges}. Such RL-based planning approaches still struggle to generalize across diverse and unseen environments \cite{generalization_gap}.

A key limitation of both rule-based and purely learning-based navigation methods is the absence of semantic understanding, which is essential for efficient goal-directed behavior in realistic environments. Object-goal navigation benefits from reasoning about object-object and object-scene relationships and from leveraging commonsense priors about given environment layouts. By integrating these semantic priors, the robotic agent can effectively constrain the search process, significantly reducing the branching factor of the search tree and prioritizing paths that align with contextual expectations. Semantic information and common sense reasoning allow the agent to infer the probable locations of goal objects even when they are not present within the agent's current field of view \cite{Chaplot2020ObjectGoal}. Recent Vision-Language Models (VLMs) such as Gemini \cite{gemini2023}, LLaVA \cite{liu2023llava}, and GPT \cite{openai2023gpt4} enable injection of commonsense and scene priors into navigation, bridging geometric perception and high-level planning \cite{ha2023semantic}. Through large-scale multimodal pretraining, these models acquire semantic priors that support generalization to previously unseen environments, reducing reliance on environment-specific training \cite{gadre2023cow, hendrycks2021natural}.

Although pretrained VLMs supply strong semantic cues, their internal heuristics, which are the learned rules that guide their behavior, are not calibrated for sequential decision making \cite{robot_ask_help}. VLM outputs are optimized for token prediction or image-text alignment rather than long-horizon action selection under physical constraints. As a result, they often assign non-trivial confidence to unsafe, redundant, or goal-irrelevant actions \cite{ahn2022as, kadavath2022language}. In this work, we tackle the problem of unreliable high-level action scoring in unseen environments. Without an explicit mechanism to quantify and control confidence, VLM-driven navigation can assign non-trivial scores to semantically plausible but suboptimal or unsafe actions. Over long horizons, such miscalibration can manifest as inefficient exploration, excessive backtracking, and action loops.

Conformal Prediction (CP) provides a principled statistical mechanism to address this issue by converting uncalibrated model scores into prediction sets\cite{vovk2005algorithmic, angelopoulos2023conformal}. Rather than trusting raw VLM confidences, CP uses a held-out calibration set to compute thresholds that control the error rate at a user-specified risk level. Under the assumption of exchangeability, this ensures that the probability of excluding the correct action is bounded by the calibration threshold, independent of the VLM architecture or training process.

\begin{figure*}[!t]
  \centering
  \includegraphics[width=1.\textwidth]{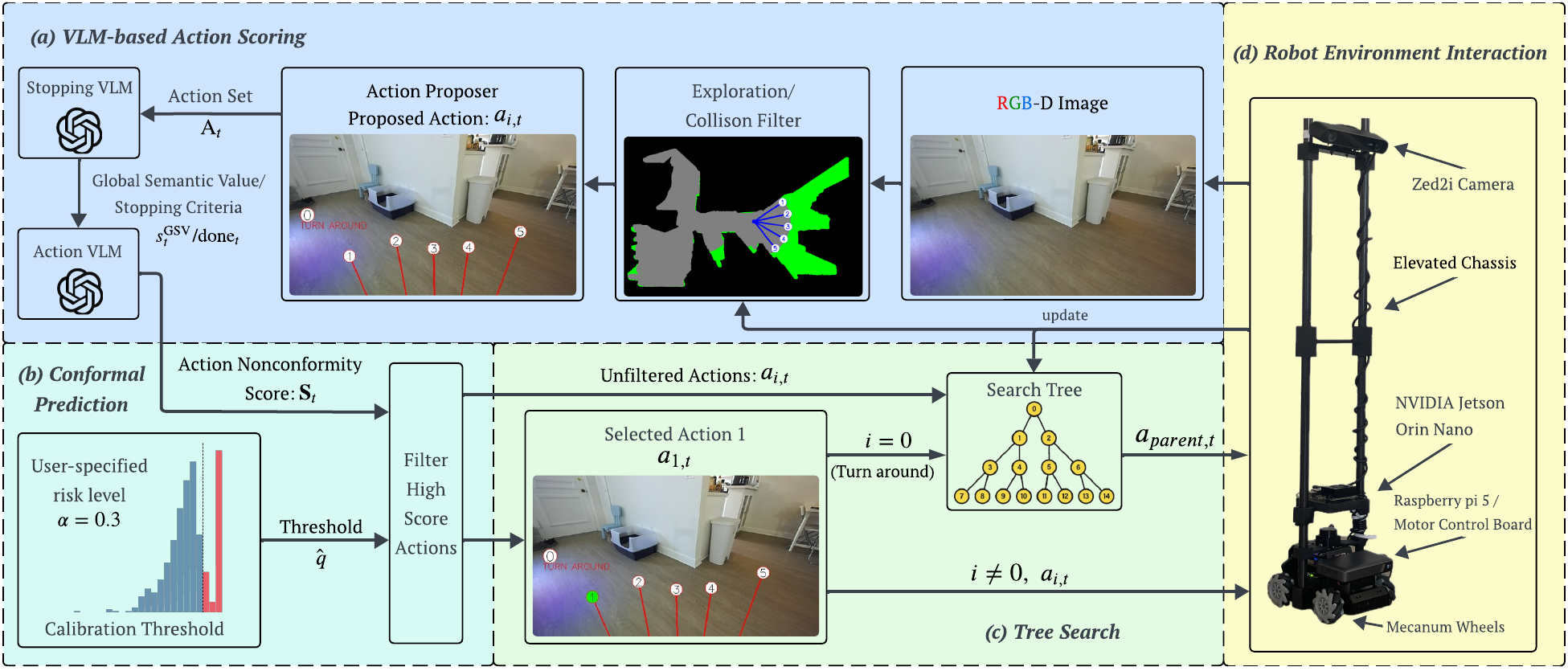}
    \caption{Overview of the \cpnav pipeline. 
    (a) The action proposer constructs candidate actions from RGB-D input, prioritizing unexplored regions while enforcing collision safety. 
    (b) The threshold obtained via offline conformal calibration is used to filter out actions whose scores fall below the calibrated bound. 
    (c) The highest-scoring untried action that satisfies the threshold is selected and executed, while the resulting transition is recorded in the exploration tree.
    (d) Hardware experiments are conducted on a modified Hiwonder MentorPi robot platform.}
  \label{fig:pipeline}
    \vspace{-5mm}
\end{figure*}

Building on this principle, we introduce \cpnav, illustrated in ~\autoref{fig:pipeline}, a conformal prediction-augmented navigation framework that integrates semantic reasoning with statistically calibrated decision-making. The main contributions of this work are as follows: (1) We introduce a conformal calibration layer that converts uncalibrated VLM action scores into statistically valid prediction sets, guaranteeing under exchangeability that an oracle-consistent trajectory is retained with probability at least $1-\alpha$, where $\alpha$ is the user-specified risk level. (2) An offline oracle-path search is formulated to identify oracle-aligned trajectories by maximizing the minimum VLM action score along a successful path, enabling trajectory-level conformal calibration for long-horizon embodied navigation. (3) Calibrated prediction sets are integrated into a tree-structured exploration policy, where calibrated VLM scores guide action selection and the stored tree supports backtracking.

\section{Related Work}
\subsection{\textbf{VLM-Augmented Vision-Language Navigation}}
Recent work leverages the zero-shot reasoning of VLMs for embodied navigation \cite{SemanticNavSurvey, gadre2023cow, ha2023semantic, chu2026abot}, covering end-to-end navigational question answering and spatial reasoning \cite{goetting2024vlmnav, shah2023lm-nav, zhou2024navgpt, zhou2024navgpt2}, hierarchical world-model and long-horizon planning \cite{wang2025tpmddn, zhang2025mem2ego, cai2025clcotnav}, task-preferenced grounding \cite{wang2025tpmddn}, and zero-shot semantic localization with frontier-based exploration \cite{gadre2023cow, Chaplot2020ObjectGoal, VLFM}. However, these methods treat VLM outputs as absolute truths, so agents are misled by hallucinated relevance or overconfident scores for inaccessible paths \cite{robot_ask_help, ahn2022as}, underscoring the need to calibrate VLM uncertainty.

\subsection{\textbf{Uncertainty Quantification for Perception and Planning}}
Uncertainty in learned perception and planning modules is commonly quantified with Bayesian neural networks (BNNs) and their approximations such as Monte Carlo dropout \cite{gal2016dropout}, deep ensembles \cite{lakshminarayanan2017simple}, or post-hoc calibration of softmax confidences \cite{gawlikowski2023survey}. However, BNNs and ensembles require access to model weights and multiple forward passes or multiple trained models, incurring substantial training and inference cost \cite{lakshminarayanan2017simple, gawlikowski2023survey}, and their confidences remain heuristic without finite-sample guarantees. These methods are also tied to a specific architecture and training procedure, so uncertainty estimation for large language and vision-language models typically falls back on sampled or verbalized outputs \cite{kadavath2022language, shorinwa2025survey}. Formal guarantees instead come from probably approximately correct (PAC) techniques such as PAC-Bayes policy bounds \cite{majumdar2021pac} and PAC confidence sets calibrated on held-out data \cite{park2020pac}, or from robust control bounding a learned perception map's error \cite{dean2020robust}, which respectively require a good prior or known linear dynamics. Conformal prediction avoids these limitations. It is model-agnostic and applies to any pretrained model without retraining, weight access, or multiple forward passes, needing only a held-out calibration set to produce prediction sets with distribution-free coverage \cite{angelopoulos2023conformal, mei2024perceive} and, with a finite-sample correction, the same PAC-style guarantee \cite{vovk2013conditional}.

\subsection{\textbf{Conformal Prediction for Reliable Decision-Making}}

Conformal prediction (CP) provides a distribution-free framework for uncertainty quantification with finite-sample coverage guarantees \cite{vovk2005algorithmic, angelopoulos2023conformal}, and several recent works have integrated it into robotics pipelines. CP has been utilized in perception and object detection to transform standard classification outputs into class prediction sets, preventing autonomous systems from making overconfident but incorrect categorical decisions \cite{mei2024perceive}. Beyond perception, CP has been applied to motion prediction and trajectory forecasting, where it constructs uncertainty-aware prediction tubes around learned dynamics models, enabling downstream planners to account for bounded future-state uncertainty \cite{lindemann2023safe, dixit2023adaptive, sheng2024safe, srinivasan2026safety}.

Existing work on high-level navigation focuses on uncertainty-aware stopping or downstream confidence estimation rather than sequential action calibration \cite{explore_until_confident, robot_ask_help, liang2024introspective}. These methods decide when to stop exploration but give no statistical guarantees for navigation actions. We apply CP directly to VLM-based action scoring, constructing calibrated prediction sets that govern the navigation policy.

\section{Method}
\subsection{\textbf{Problem Definition}}


We consider object-goal navigation in previously unseen indoor environments. An embodied agent is initialized at an unknown pose, $\mathbf{x}_0 = (x_0, y_0, \theta_0)$, where $(x_0,y_0)\in\mathbb{R}^2$ denotes the planar position and $\theta_0\in [-\pi, \pi]$ denotes the robot orientation. Given a target object category $g$, such as chair, bed, or toilet, the goal is to navigate to any visible instance of that category without prior knowledge of the map or goal location. At each time step $t$, the agent receives an RGB-D observation $I_t = (I_t^{\text{RGB}}, I_t^{\text{depth}})$ and its pose $\mathbf{x}_t = (x_t, y_t, \theta_t)$. Based on the current observation history, the agent selects a motion command $a_t = (\phi_t, d_t)$, where $\phi_t$ denotes the commanded heading direction relative to the robot's current pose and $d_t$ denotes the commanded travel distance.

\subsection{\textbf{VLM-based Action Scoring}}
\cpnav proposes collision-free candidate actions from RGB-D input and queries two VLMs: a stopping VLM that decides goal-reaching and computes a global semantic value, and an action VLM that scores each candidate.
\subsubsection{\textbf{Action proposer}}
\begin{figure}[ht]
  \centering
  \includegraphics[width=0.48\textwidth]{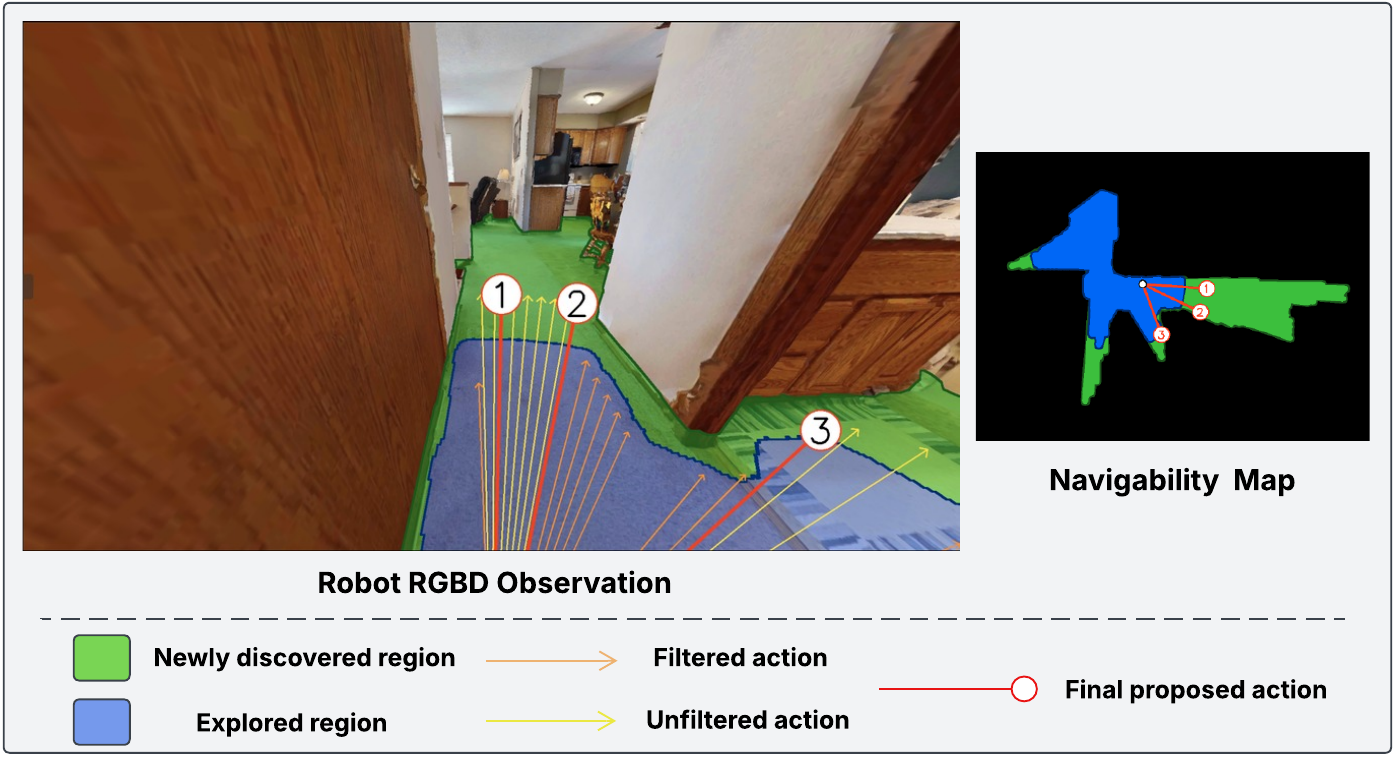}
\caption{The action proposer uses the RGB-D observation to identify newly discovered regions and generate collision-free candidate actions toward them.}
  \label{fig:action_propose}
    \vspace{-2mm}
\end{figure}

Building upon VLMNav~\cite{goetting2024vlmnav}, the action proposer generates collision-aware motion primitives from the current RGB-D observation (\autoref{fig:pipeline}a). The robot first builds a navigability mask from the depth image $I_t^{\text{depth}}$, treating regions with estimated height variation below $5\,\mathrm{cm}$ as traversable (\autoref{fig:action_propose}). It then samples evenly spaced ray directions $\phi_{i,t}$ across its horizontal field of view, where $i$ indexes each direction, and computes the maximum collision-free travel distance $d_{i,t} = \max \{ d \in [0,d_{\max}] \mid \phi = \phi_{i,t} \}$, where $d_{\max}$ is the maximum considered travel distance, using a footprint inflated by $r_{\text{robot}} + r_{\text{clearance}}$ to account for localization and depth errors.

To prioritize efficient exploration, we remove primitives shorter than 0.75m and clip longer rays to 1.7m, so that long actions do not miss local observations. We also remove actions whose endpoints lie mostly in already explored regions. Finally, we include an auxiliary turn-around action $a_{0,t} = (\pi, 0)$ as a fallback when no forward candidate is selected. The final action set is $\mathrm{A}_t = \{ a_{i,t}=(\phi_{i,t}, d_{i,t}) \}_{i=0}^{N}$, where $N$ is the number of primitives remaining after filtering.

\subsubsection{\textbf{VLM Scoring and Stopping}}
\begin{figure}[ht]
  \centering
  \includegraphics[width=0.475\textwidth]{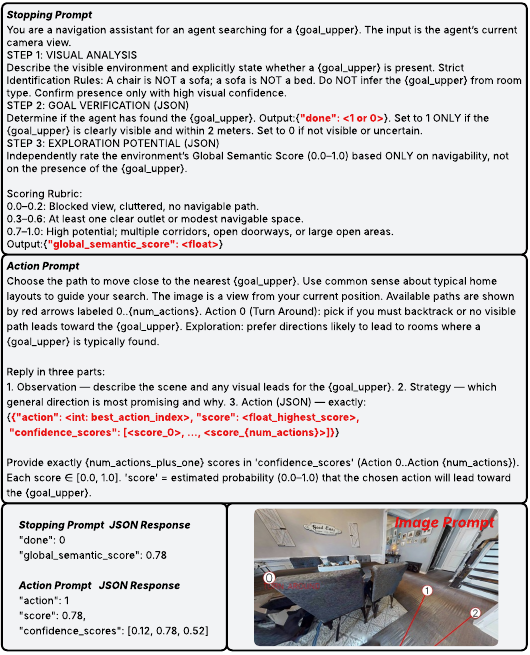}
  \caption{Concise stopping and action prompts with their JSON responses. The stopping prompt,  \( f_{\text{stop}}(\cdot) \) queries the stopping condition, \( \text{done}_t \), and GSV, $s_t^{\text{GSV}}$; the action prompt,  \( f_{\text{action}}(\cdot) \), queries the per-action confidence score, $s_{i,t}^{\text{action}}$. The manually normalized confidence score is multiplied by GSV to form the final per-action score, ${s}_{i,t}$.}
  \label{fig:vlm}
    \vspace{-5mm}
\end{figure}

Given the current RGB observation $I_t^{\text{RGB}}$ and the object-goal description $g$, the stopping VLM outputs,
  \vspace{-1mm}
  \begin{equation}
(\text{done}_t,\, s_t^{\text{GSV}}) = f_{\text{stop}}(I_t^{\text{RGB}}, g),
  \vspace{-1mm}
\end{equation}
where $f_{\text{stop}}(\cdot)$ denotes the stopping VLM query, $\text{done}_t \in \{0,1\}$ indicates whether the agent sees the goal, and $s_t^{\text{GSV}} \in [0,1]$ is the global semantic value (GSV), which measures how promising the current view is for exploration. For each candidate action $a_{i,t} \in \mathrm{A}_t$, the action VLM assigns a semantic confidence score,
 \vspace{-1mm}
\begin{equation}
s_{i,t}^{\text{action}} = f_{\text{action}}(I_t^{\text{RGB}}, a_{i,t}, g),
 \vspace{-1mm}
\end{equation}
where $f_{\text{action}}(\cdot)$ denotes the action VLM query and $s_{i,t}^{\text{action}} \in [0,1]$, with higher values indicating a more promising action. To incorporate scene-level navigation potential, the two are fused into the per-action score $s_{i,t} = s_t^{\text{GSV}} \cdot s_{i,t}^{\text{action}}$, and the confidence vector $\mathbf{S}_t = [s_{0,t}, \dots, s_{N,t}]$ over all $N+1$ candidate actions in $\mathrm{A}_t$ is used for decision-making at each step. Concise versions of the stopping and action prompts with their JSON responses are shown in \autoref{fig:vlm}.

\subsection{\textbf{Tree-Structured Exploration with Backtracking}}
We represent the agent's exploration history as a dynamically growing tree to support action selection and backtracking. Each node records a visited state, its candidate actions, and their VLM scores, while each edge represents an executed action. The tree is an exploration-history structure rather than a planning tree: the policy performs no rollouts, cost optimization, or long-horizon value estimation.
\subsubsection{\textbf{Exploration Tree Representation}}
Let $k$ denote the current time step. The robot maintains an exploration log $\mathrm{H}_k = \{(\mathbf{x}_t, \mathrm{A}_t, \mathbf{S}_t)\}_{t=1}^{k}$, which stores the state, candidate actions, and fused scores of every visited step $t \le k$. The parent of node $k$ is the node $t$ from which the action reaching $\mathbf{x}_k$ was executed, and $\mathrm{H}_k$ stores the associated data.

\subsubsection{\textbf{Local Backtrack}}

If the VLM selects the turn-around action (index 0), the robot reverts to the parent node $t^\ast $ of the current time step $k$ and retrieves its stored state and action set. The remaining untried actions are $\mathrm{R}_{t^\ast} = \{ a_{i,t^\ast} \in \mathrm{A}_{t^\ast} \mid a_{i,t^\ast} \notin \mathrm{T}_{t^\ast} \}$, where $\mathrm{T}_{t} \subseteq \mathrm{A}_{t}$ denotes the set of actions already attempted from step~$t$. 

If $\mathrm{R}_{t^\ast}$ is nonempty, the remaining actions are sorted by their semantic confidence scores and the highest-ranked untried action is executed from the parent state $\mathbf{x}_{t^\ast}$. Otherwise, the agent initiates a global backtrack. At the root node, the turn-around action is executed directly.

\subsubsection{\textbf{Global Backtrack}}

If local backtrack exhausts all recent branches, a global backtrack searches the full history up to the current time step $k$ for any node that still contains untried actions,
 \vspace{-1mm}
\begin{equation}
\mathrm{R}^{\mathrm{global}}_k
=
\bigcup_{t=1}^{k}
\left\{
a_{i,t} \in \mathrm{A}_t
\;\middle|\;
a_{i,t} \notin \mathrm{T}_t,\; i \neq 0
\right\}.
 \vspace{-1mm}
\end{equation}
If $\mathrm{R}^{\mathrm{global}}_k$ is empty, exploration terminates.
Otherwise, the agent is restored to the state $\mathbf{x}_t$ of the highest-scoring action in $\mathrm{R}^{\mathrm{global}}_k$ and executes it.

\subsection{\textbf{Offline Oracle-Path Search}}

During offline calibration, once the goal is first reached, we build a candidate path to the goal. However, this path may be circuitous or include low-confidence actions. To find the highest-confidence goal-reaching path and avoid overly conservative calibration, we explore the remaining untried actions. In the exploration tree, nodes are visited states $\mathbf{x}_t$ and edges are executed actions $a_{i,t}$, each carrying its fused score $s_{i,t}$. Let $P$ denote a path from the root to a goal-reaching node. The oracle path $P^*$ is the path for which the overall path score is maximized,
 \vspace{-1mm}
\begin{equation}
P^* = \arg\max_{P} \,\, b(P) = \arg\max_{P} \left( \min_{a_{i,t} \in P} s_{i,t} \right).
 \vspace{-1mm}
\end{equation}
We call the minimum score along a path its bottleneck, denoted $b(P)$. The bottleneck $b(P^*)$ of the oracle path uses a trajectory-level confidence for calibration instead of a step-wise score. This is because during calibration, we want to ensure the  calibration is performed over the entire trajectory and the prediction set retains a trajectory exactly when it retains its lowest-scoring action. We implement this with a modified Breadth-First Search that prioritizes paths with higher bottleneck scores.

To further optimize the oracle path, the agent re-enters a global backtracking phase restricted to untried actions with scores strictly exceeding the current bottleneck,
\begin{equation}
\mathrm{R}^{\mathrm{refine}}_k = \{ a_{i,t} \in \mathrm{R}^{\mathrm{global}}_k \mid s_{i,t} > b(P^*) \}.
\end{equation}
The search continues until the agent reaches the goal again, triggers a turn-around, or converges to an existing node, and is repeated until the total step limit is met. If a higher bottleneck score is found, $P^*$ is updated, so the final threshold used for conformal prediction represents the most reliable path discovered.

\subsection{\textbf{Offline Calibration with Conformal Prediction}}

\subsubsection{\textbf{Non-conformity Scoring}}

The calibration set consists of $n = 500$ independent and identically distributed episodes sampled with a fixed random seed, each containing a randomly sampled scene, initial pose, object goal, and the corresponding oracle-action trajectory. For each episode $j$, the trajectory-level non-conformity score is the minimum oracle-action score along the oracle path, $E_j = 1-\min_t s_{a^*,t,j} = 1-b(P_j^*)$, where $s_{a^*,t,j}$ is the VLM score of the oracle action $a^*$ at step $t$ of trajectory $j$. The VLM gives unnormalized scores, which reduces bias, and we normalize them for cross-step consistency.

\subsubsection{\textbf{Dataset-Conditional Uncertainty Alignment}}
Because the calibration set is finite, the empirical quantile alone may not guarantee the target coverage $1-\alpha$ on new data. We therefore use a failure probability $\delta=0.01$ and define a stricter adjusted risk level $\hat{\alpha}$ such that,
\begin{equation}~\label{eq:beta}
\operatorname{Beta}^{-1}\left(\delta;\,n+1-v,\,v\right)\geq 1-\alpha,
\end{equation}
where $v=\lfloor(n+1)\hat{\alpha}\rfloor$ and $\operatorname{Beta}^{-1}(\delta;a,b)$ is the $\delta$-quantile of the $\operatorname{Beta}(a,b)$ distribution. This ensures, with confidence $1-\delta$ over the calibration set, that the oracle-consistent trajectory is retained with probability at least $1-\alpha$ on a new exchangeable episode.

\subsubsection{\textbf{Quantile Application}}
Using the adjusted risk $\hat{\alpha}$, we compute the calibrated threshold $\hat{q}$ as the conformal quantile of the non-conformity scores \cite{vovk2005algorithmic, angelopoulos2023conformal},
\begin{equation}
\hat{q} = \text{Quantile}\left(\{E_1, \dots, E_n\}, \frac{\lceil (n+1)(1-\hat{\alpha}) \rceil}{n}\right).
\end{equation}

The calibrated quantile $\hat q$ is then used to construct the admissible prediction set during online deployment.

\subsection{\textbf{Online Deployment with Calibrated Prediction Sets}}

\subsubsection{\textbf{Prediction Set Construction}}
At each time step $t$, the agent constructs a prediction set $C(\xi_t)$, where $\xi_t = (I_t^{\text{RGB}}, g, \mathrm{A}_t)$ denotes the query context at step $t$, containing the candidate actions that satisfy the calibrated threshold,
\begin{equation}
C(\xi_{t}) = \{ a_{i,t} \in \mathrm{A}_t \mid s_{i,t} \ge 1 - \hat{q} \}
\label{eq:prediction_set}
\end{equation}
If $C(\xi_t) = \emptyset$, the agent triggers the turn-around action $a_{0,t}$ to initiate backtracking. Otherwise, it executes the highest-scoring action in $C(\xi_t)$.

Let $a_t^*$ denote the oracle action at time $t$. Under exchangeability between calibration and deployment episodes, the conformal threshold $\hat q$ ensures,
 \vspace{-1mm}
\begin{equation}
\mathbb{P}\!\left(a_t^* \in C(\xi_t), \forall t>0 \right) \ge 1-\alpha.
 \vspace{-1mm}
\end{equation}

This guarantee holds at the trajectory level: with probability at least $1-\alpha$, the oracle-consistent path is not pruned during online exploration. The correction~(\ref{eq:beta}) further ensures that this coverage bound holds with confidence $1-\delta=0.99$ over the sampling of the finite calibration set. The proof is given in Appendix~\ref{app:proof}.

\subsubsection{\textbf{Score-Guided Tree Exploration}}
Among the admissible actions, the agent selects the highest-scoring untried action and records the resulting transition in the exploration tree. If the turn-around action is selected, or all admissible actions along the current branch are exhausted, the agent performs the local and global backtracking procedures of Sec.~III-C.

\section{Experimental Setup}

\subsection{\textbf{Simulation and Hardware Configuration}}

In simulation, we evaluate \cpnav{} on the Habitat-Sim platform using HM3D v0.2 as the primary benchmark and MP3D for cross-dataset generalizability \cite{Habitat, HM3D, chang2017matterport3d}. For hardware, calibration uses a simulator configuration that reflects the physical platform.

The hardware platform is a Hiwonder MentorPi M1~\cite{hiwonder_mentorpi} with a Mecanum drivetrain. State estimation uses the SLAM Toolbox, fusing a 2D LiDAR, wheel odometry, and an IMU, and a ZED 2i stereo camera~\cite{stereolabs_zed2i} provides rectified RGB-D data. Table~\ref{tab:exp_params} lists the experimental parameters.

\begin{table}[h]
\centering
\caption{Experimental Parameters}
\label{tab:exp_params}
\begin{tabular}{l|cc}
\hline
\textbf{Parameter} & \textbf{Simulation} & \textbf{Hardware} \\ \hline
$r_{\mathrm{robot}},\; r_{\mathrm{clearance}}$
    & \multicolumn{2}{c}{$0.1\,\mathrm{m},\; 0.05\,\mathrm{m}$} \\
Resolution
    & \multicolumn{2}{c}{$640 \times 480$ RGB-D} \\
Camera Height
    & \multicolumn{2}{c}{$0.9\,\mathrm{m}$} \\
HFoV / Down Pitch
    & $131^{\circ} / 0.45\,\mathrm{rad}$
    & $101^{\circ} / 0.4\,\mathrm{rad}$ \\
VLM Model
    & \multicolumn{2}{c}{GPT-5 Nano~\cite{openai2025gpt5}} \\
Step Limit
    & 200
    & 100 \\
Target Risk ($\alpha$)
    & 0.3
    & 0.45 \\
Failure Probability ($\delta$)
    & \multicolumn{2}{c}{0.01} \\
Episodes (Cal/Eval)
    & 500 / 200
    & 500 / 24 (Real-world) \\
\hline
\end{tabular}
\end{table}
\subsection{\textbf{Baselines and Metrics}}

We compare \cpnav with alternative strategies that incorporate uncertainty in different ways~\cite{robot_ask_help}, as well as with STRIVE~\cite{zhu2025strive}, the state-of-the-art object-navigation method, and WMNav~\cite{nie2025wmnav}, which uses macro-actions similar to those in the \cpnav pipeline.

\begin{itemize}
    \item \textbf{No Calibration:} The agent executes the action with the
    highest raw VLM score $s_{i,t}$ without uncertainty calibration,  following the VLMNav baseline~\cite{goetting2024vlmnav};

    \item \textbf{Simple Set:} Actions are ranked by score and added in
    descending order to form the smallest set whose cumulative score reaches
    at least $1-\alpha$;
    
    \item \textbf{Prompt Set:} The VLM is prompted to directly list all
    feasible actions that could lead to the goal;

    \item \textbf{Ensemble Set:} Builds on the Simple Set by querying the VLM
    multiple times and using response frequencies to construct the confidence
    vector $\mathbf{S}_t$.

\end{itemize}

All methods are evaluated on the same 200 episodes per benchmark. We grade all methods with the same manual criterion: an episode succeeds only when the correct target is visible in the final observation and reachable through an unobstructed path, and stopping near an incorrect target counts as failure. We report Success Rate (SR), Success weighted by Path Length (SPL)~\cite{SPL_metric} using the Habitat geodesic distance to the verified goal, and mean distance traveled. On hardware, we compare \cpnav against the No Calibration baseline over 24 trials each in two apartment environments with floor plans shown in Appendix~\ref{app:hw_env}, with random initial poses and goals drawn from the simulation goal set, using the same metrics.

\section{RESULTS}
\subsection{\textbf{Simulation Results}}
\subsubsection{\textbf{Offline Calibration Threshold Analysis}}

\begin{figure}[t]
  \vspace{-2mm}
  \centering
  \includegraphics[width=0.49\textwidth]{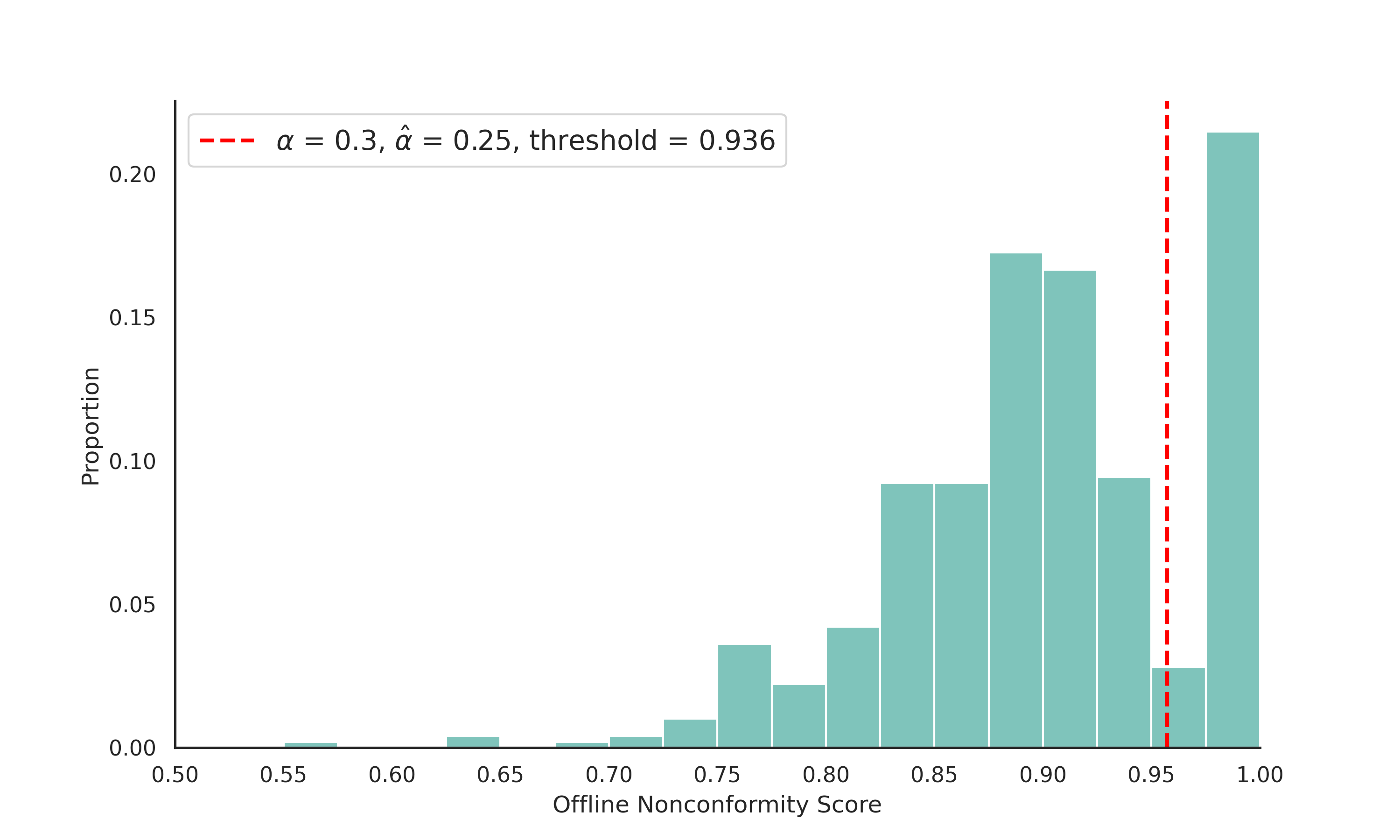}
  \caption{Empirical distribution of nonconformity scores \(E_j\) used for simulation conformal calibration. The dashed red line indicates the calibrated nonconformity quantile \(\hat q = 0.940\), evaluated at the Beta-adjusted quantile level \(1-\hat{\alpha}\).}
  \label{fig:calibration_histograms}
  \vspace{-7mm}
\end{figure}

\autoref{fig:calibration_histograms} shows the empirical distribution of oracle-action nonconformity scores used for calibration, with the dashed red line marking the calibrated threshold $\hat q$. With the correction at $\delta=0.01$, the simulation setting uses a target risk of $\alpha=0.3$ and yields $\hat{q}_{\mathrm{sim}} = 0.940$ at an adjusted risk level of $\hat{\alpha}=0.25$. The hardware setting uses $\alpha=0.45$ and yields $\hat{q}_{\mathrm{real}} = 0.936$ at $\hat{\alpha}=0.39$. Because larger $E_j$ corresponds to lower confidence, $\hat q$ lies in the upper tail of the distribution, so conformal prediction removes only the lowest-confidence fraction of oracle-aligned actions while retaining the majority of reliable decisions.

\subsubsection{\textbf{Simulation Results}}

\begin{table}[t]
\centering
\caption{Simulation evaluation on HM3D and MP3D
(200 episodes per benchmark). Distance is reported in meters.}
\label{tab:sim_results}
\renewcommand{\arraystretch}{1.08}
\setlength{\tabcolsep}{4pt}
\resizebox{\columnwidth}{!}{%
\begin{tabular}{@{}l|ccc|ccc@{}}
\toprule
\textbf{Policy}
& \multicolumn{3}{c|}{\textbf{HM3D}}
& \multicolumn{3}{c}{\textbf{MP3D}} \\
\cmidrule(lr){2-4}\cmidrule(lr){5-7}
& \textbf{SR (\%)} $\uparrow$
& \textbf{SPL (\%)} $\uparrow$
& \textbf{Dist. (m)}
& \textbf{SR (\%)} $\uparrow$
& \textbf{SPL (\%)} $\uparrow$
& \textbf{Dist. (m)} \\
\midrule
No Calibration~\cite{goetting2024vlmnav}
& 81.0 & 40.3 & 35.4
& 68.5 & 43.0 & 36.4 \\

Simple Set
& 76.7 & 39.5 & 32.8
& 71.0 & 44.3 & 33.4 \\

Prompt Set
& 79.7 & 38.7 & 37.4
& 71.5 & 45.3 & 31.6 \\

Ensemble Set
& 74.0 & 39.1 & 28.0
& 66.5 & 41.8 & 28.2 \\

STRIVE~\cite{zhu2025strive}
& 82.5 & 41.5 & \textbf{17.6}
& 71.5 & 42.3 & 21.0 \\

WMNav~\cite{nie2025wmnav}
& 80.0 & 39.9 & 18.2
& 75.0 & 45.2 & \textbf{18.5} \\

\midrule
\textbf{\cpnav{} (Ours)}
& \textbf{86.0}
& \textbf{44.9}
& 29.2
& \textbf{77.0}
& \textbf{46.8}
& 32.1 \\
\bottomrule
\end{tabular}%
}
\end{table}

Table~\ref{tab:sim_results} reports the simulation results for $\hat q_{\text{sim}} = 0.94$. Among the calibration baselines, \cpnav achieves the highest SR of $86.0\%$. Relative to No Calibration, it improves SR by $6.2\%$ and SPL by $11.4\%$ while reducing mean travel distance by $17.5\%$, and it outperforms the best heuristic threshold-based variant by $12.1\%$ in relative SR and $13.7\%$ in relative SPL. Because every episode is capped, steps spent on low-confidence branches cannot be recovered later, so pruning them raises the fraction of episodes that reach the goal before the cap and improves SR and SPL together.

Under each method's original step limit, \cpnav slightly improves SR and SPL over STRIVE and WMNav, although their different action representations and step definitions make step limits difficult to normalize. STRIVE travels $39.7\%$ and $34.6\%$ shorter mean distances than \cpnav on HM3D and MP3D, and WMNav $37.7\%$ and $42.4\%$ shorter. The longer distance of \cpnav reflects the complete backtracking required by its goal-reaching guarantee, which adds travel distance when the goal lies in a lower-ranked region, so \cpnav prioritizes search coverage and reliability over direct travel efficiency. Appendix~\ref{app:step_budget} reports \cpnav under reduced step budgets, where it retains its advantage over the uncalibrated baseline, and Appendix~\ref{app:tokens} compares the token consumption of all methods. We also emphasize that the guarantees provided by our conformal prediction-based framework, \cpnav, can be applied to any embodied navigation framework like WMNav or STRIVE.

\subsection{\textbf{Simulation Parameter Sensitivity Analysis}}

We vary one parameter at a time from the default \cpnav setting (GPT-5o-nano, height 0.9\,m, HFoV 131$^\circ$, $\hat q = 0.940$, voxel map enabled). Table~\ref{tab:param_sensitivity} reports SR, SPL, and mean travel distance.

\begin{table}[h]
\centering
\caption{Parameter sensitivity analysis in simulation.}
\label{tab:param_sensitivity}
\setlength{\tabcolsep}{4pt}
\begin{tabular}{lccc}
\toprule
\textbf{Variant} & \textbf{SR (\%)} & \textbf{SPL (\%)} & \textbf{Mean Distance (m)} \\
\midrule
\textbf{\cpnav ($\hat q = 0.940$)} & \textbf{86.0} & \textbf{44.9} & \textbf{29.2} \\
No Calibration & 81.0 & 40.3 & 35.4 \\
\midrule
Height = 0.45 m & 67.0 & 31.4 & 44.6 \\
Height = 1.70 m & 79.3 & 40.5 & 30.4 \\
\midrule
HFoV = 69$^\circ$ & 40.0 & 18.7 & 44.8 \\
HFoV = 101$^\circ$ & 72.7 & 30.5 & 40.9 \\
\midrule
Model = GPT-4o-mini~\cite{openai2023gpt4} & 71.3 & 33.7 & 34.4 \\
\midrule
No Voxel Map & 66.0 & 36.4 & 36.5 \\
\bottomrule
  \vspace{-5mm}
\end{tabular}
\end{table}
The table highlights three main trends. First, sensing geometry strongly affects performance, as lower camera height or narrower field of view reduces SR and SPL while increasing travel distance. Second, model capacity matters, since replacing GPT-5o-nano with a smaller VLM degrades both reliability and efficiency. Finally, the voxel-based occupancy map improves exploration by reducing redundant revisits to different states.


\subsection{\textbf{Hardware Results}}

We evaluated \cpnav across 24 randomized object-goal episodes in two apartment layouts with $\hat{q} = 0.936$, corresponding to $\alpha=0.45$. Table~\ref{tab:hw_results} compares it with the uncalibrated baseline.

\begin{table}[h]
\centering
\caption{Hardware results.}
\label{tab:hw_results}
\begin{tabular}{lccc}
\hline
\textbf{Policy} & \textbf{SR (\%)} & \textbf{SPL (\%)} & \textbf{Mean Distance (m)} \\
\hline
No Calibration & 66.7 & 42.0 & 9.0 \\
\textbf{\cpnav} & \textbf{83.3} & \textbf{48.4} & \textbf{6.3} \\
\hline
\end{tabular}
\end{table}

SR increased from 66.7\% to 83.3\%, a relative improvement of $24.9\%$, SPL increased by $15.2\%$, and mean distance decreased by $30.0\%$. These gains exceed those in simulation, since the reduced maximum step count and execution noise amplify the cost of the suboptimal actions that calibration filters out.Differences between simulation and hardware stem from the reduced HFoV ($101^\circ$ versus $131^\circ$), state-estimation drift during backtracking, and the smaller apartments, which explain the higher hardware success rates.

\section{Conclusion and Limitations}

\textbf{Conclusions.}
We proposed \cpnav, a conformal prediction-based calibration framework for vision-language navigation. By filtering low-confidence actions through conformal calibration, \cpnav improves reliability and path efficiency without modifying the underlying policy, achieving higher SR and SPL in simulation and on hardware, where strict step budgets and execution noise amplify the cost of suboptimal decisions. The calibration scores also show how well a VLM's action scores align with successful paths. The layer uses only output scores, so it applies to any VLM-based navigation framework without retraining.


\textbf{Limitations.}
First, tree nodes are defined solely by spatial proximity without orientation, which may introduce poor estimation. Second, hardware calibration uses simulated episodes, so the guarantee relies on exchangeability between simulation and the real apartments. Finally, the method relies on scores from a VLM accessed via the ChatGPT API, whose stochasticity can introduce variability in action scoring and calibration. Future work will explore orientation-aware states and robustness to VLM variability.





\bibliographystyle{IEEEtran}
\bibliography{refs}

\clearpage
\clearpage
\appendix
\subsection{Proof of the High-Probability Goal-Reaching Guarantee}
\label{app:proof}

\phantomsection
\label{app:goal_proof}

\noindent\textbf{Proposition 1.}
Suppose a goal-reaching oracle trajectory exists within the candidate action structure. Under exchangeability between the calibration and test episodes, complete backtracking over the actions retained by the conformal prediction sets, reliable action execution, and a sufficient step budget, \cpnav reaches the goal with probability at least $1-\alpha$.

\medskip
\noindent\textit{Proof.}
For calibration trajectory $j$, define the trajectory-level
nonconformity score as
\[
E_j = 1-\min_t s_{a^*,t,j}
\]
For a new test episode, define
\[
E_{\mathrm{test}}
=
1-\min_t s_{a_t^*,t}^{\mathrm{test}}.
\]
By the conformal coverage guarantee,
\[
\mathbb{P}
\left(
E_{\mathrm{test}}\leq \hat q
\right)
\geq 1-\alpha.
\]

Whenever $E_{\mathrm{test}}\leq\hat q$,
\[
1-\min_t s_{a_t^*,t}^{\mathrm{test}}
\leq \hat q,
\]
which implies
\[
s_{a_t^*,t}^{\mathrm{test}}
\geq 1-\hat q,
\qquad \forall t.
\]
By the definition of the conformal prediction set,
\[
C(\xi_t)
=
\left\{
a_{i,t}\in \mathrm{A}_t
\mid
s_{i,t}\geq 1-\hat q
\right\},
\]
every oracle action satisfies
\[
a_t^*\in C(\xi_t),
\qquad \forall t.
\]
Therefore, with probability at least $1-\alpha$, the complete
goal-reaching oracle trajectory is not removed by conformal pruning.

The policy may initially execute other admissible actions ranked by their score. Assuming there exists a path to the goal and the step is finite, complete backtracking and a sufficient step budget ensure that every retained untried action is eventually explored. Since all actions in the goal-reaching oracle trajectory $P^*$ remain admissible, the policy eventually traverses $P^*$. Since the retained oracle trajectory $P^*$ is eventually explored,
\[
\mathbb{P}\!\left(\cpnav\text{ reaches the goal}\right)
\geq
\mathbb{P}\!\left(E_{\mathrm{test}}\leq\hat q\right)
\geq 1-\alpha.
\]

\subsection{Hardware Evaluation Environments}
\label{app:hw_env}

\begin{figure}[H]
    \centering
    \includegraphics[width=0.46\columnwidth]{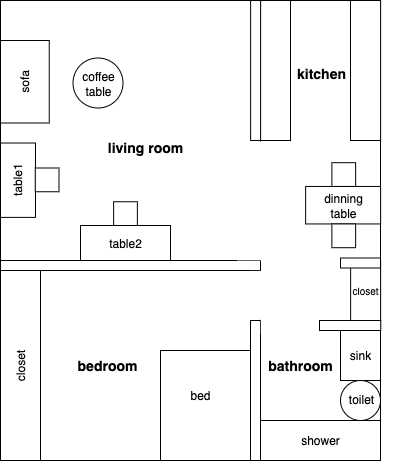}%
    \hspace{0.02\columnwidth}%
    \includegraphics[width=0.46\columnwidth]{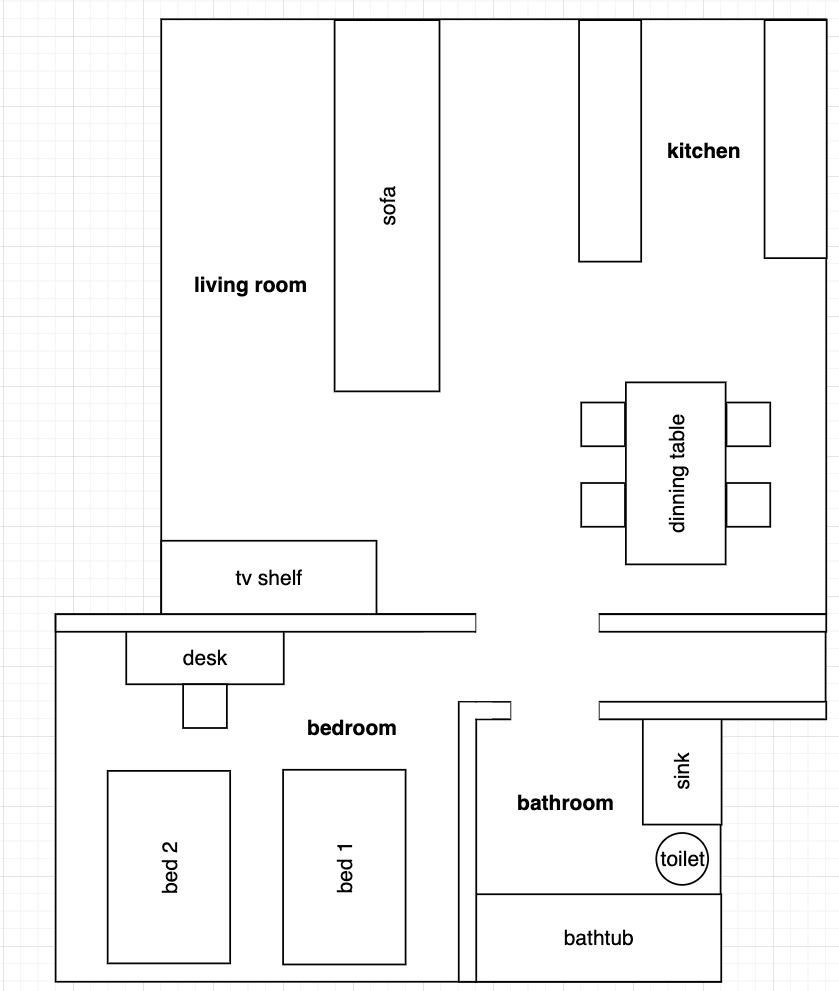}
    \caption{Floor plans of the two hardware evaluation environments.}
    \label{fig:hardware_floorplans}
\end{figure}

Figure~\ref{fig:hardware_floorplans} shows the two 1B1B environments used for hardware evaluation. Both are standard apartment settings and are considered in-distribution with respect to the residential environments represented in HM3D and MP3D.

\subsection{Performance under Restricted Step Budgets}
\label{app:step_budget}

We evaluate \cpnav on HM3D v0.2 at 50, 100, and 200 steps, corresponding to 25\%, 50\%, and 100\% of the full budget.

\begin{table}[H]
\centering
\setlength{\tabcolsep}{3pt}
\caption{Performance under restricted step budgets.}
\label{tab:app_hm3d_step_budget}
\resizebox{\columnwidth}{!}{%
\begin{tabular}{@{}lcccccc@{}}
\toprule
\textbf{Policy}
& \multicolumn{2}{c}{\textbf{50 Steps (25\%)}}
& \multicolumn{2}{c}{\textbf{100 Steps (50\%)}}
& \multicolumn{2}{c}{\textbf{200 Steps (100\%)}} \\
\cmidrule(lr){2-3}\cmidrule(lr){4-5}\cmidrule(lr){6-7}
& \textbf{SR (\%)} & \textbf{SPL (\%)}
& \textbf{SR (\%)} & \textbf{SPL (\%)}
& \textbf{SR (\%)} & \textbf{SPL (\%)} \\
\midrule
No Calibration
& 60.3 & 37.7
& 75.7 & 39.8
& 81.0 & 40.3 \\
\textbf{\cpnav}
& \textbf{73.0} & \textbf{43.2}
& \textbf{80.0} & \textbf{44.4}
& \textbf{86.0} & \textbf{44.9} \\
\bottomrule
\end{tabular}%
}
\end{table}

Relative to No Calibration, \cpnav improves SR by 21.1\%, 5.7\%, and 6.2\% and SPL by 14.6\%, 11.6\%, and 11.4\% at 50, 100, and 200 steps, demonstrating stronger performance under limited budgets.


\begin{figure*}[t]
    \centering
    \setlength{\fboxsep}{0pt}
    \begin{subfigure}[t]{0.48\textwidth}
        \centering
        \colorbox{white}{%
            \parbox[c][0.9504\linewidth][t]{\linewidth}{%
                \centering
                \includegraphics[trim=0 237bp 0 0,clip,width=\linewidth]{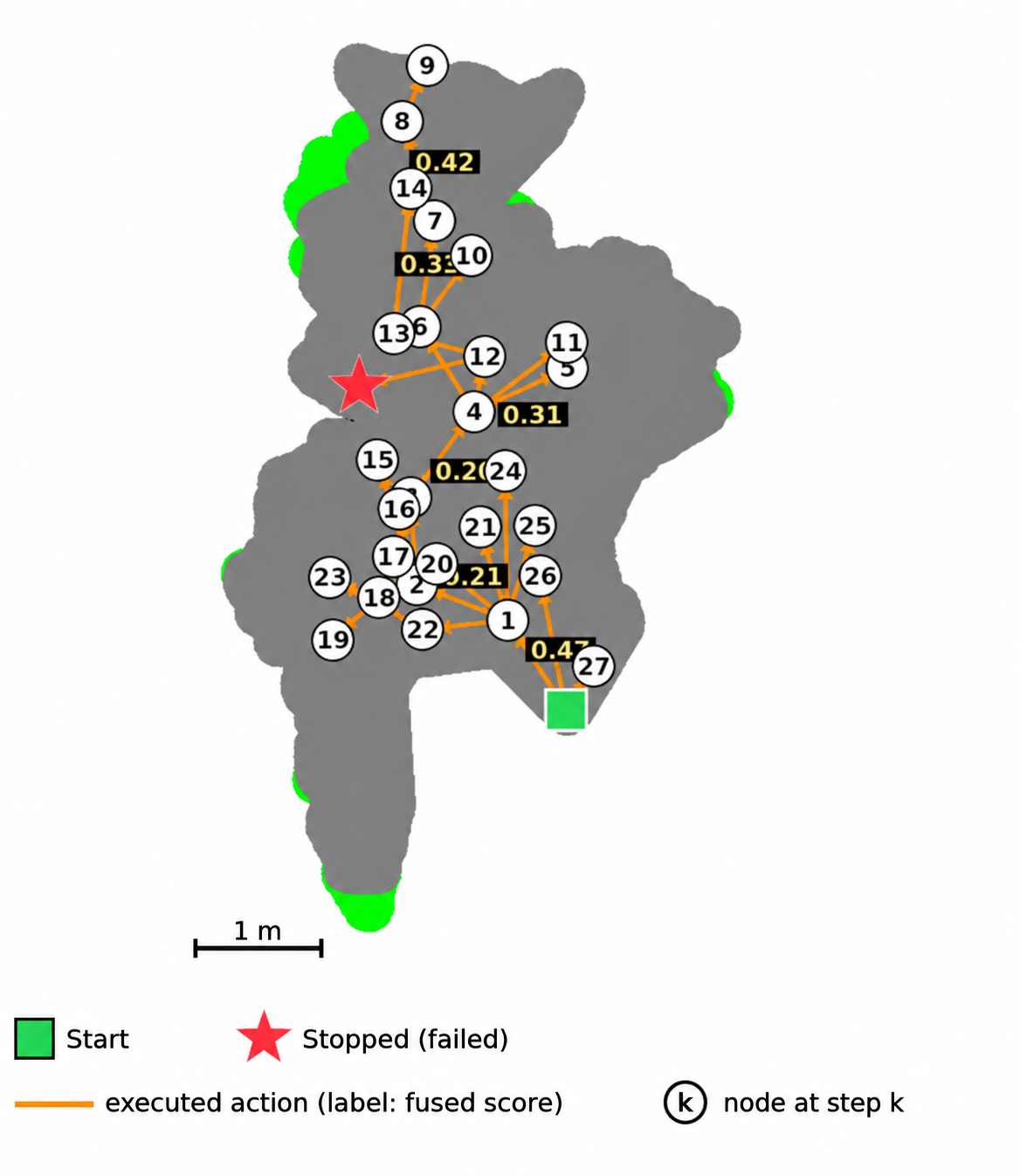}%
            }%
        }
        \caption{No Calibration: failure after 44 steps.}
        \label{fig:hm3d_topdown_baseline}
    \end{subfigure}
    \hfill
    \begin{subfigure}[t]{0.48\textwidth}
        \centering
        \colorbox{white}{%
            \parbox[c][0.9504\linewidth][t]{\linewidth}{%
                \centering
                \vspace*{0.0746\linewidth}
                \includegraphics[trim=0 196bp 0 0,clip,width=0.86878\linewidth]{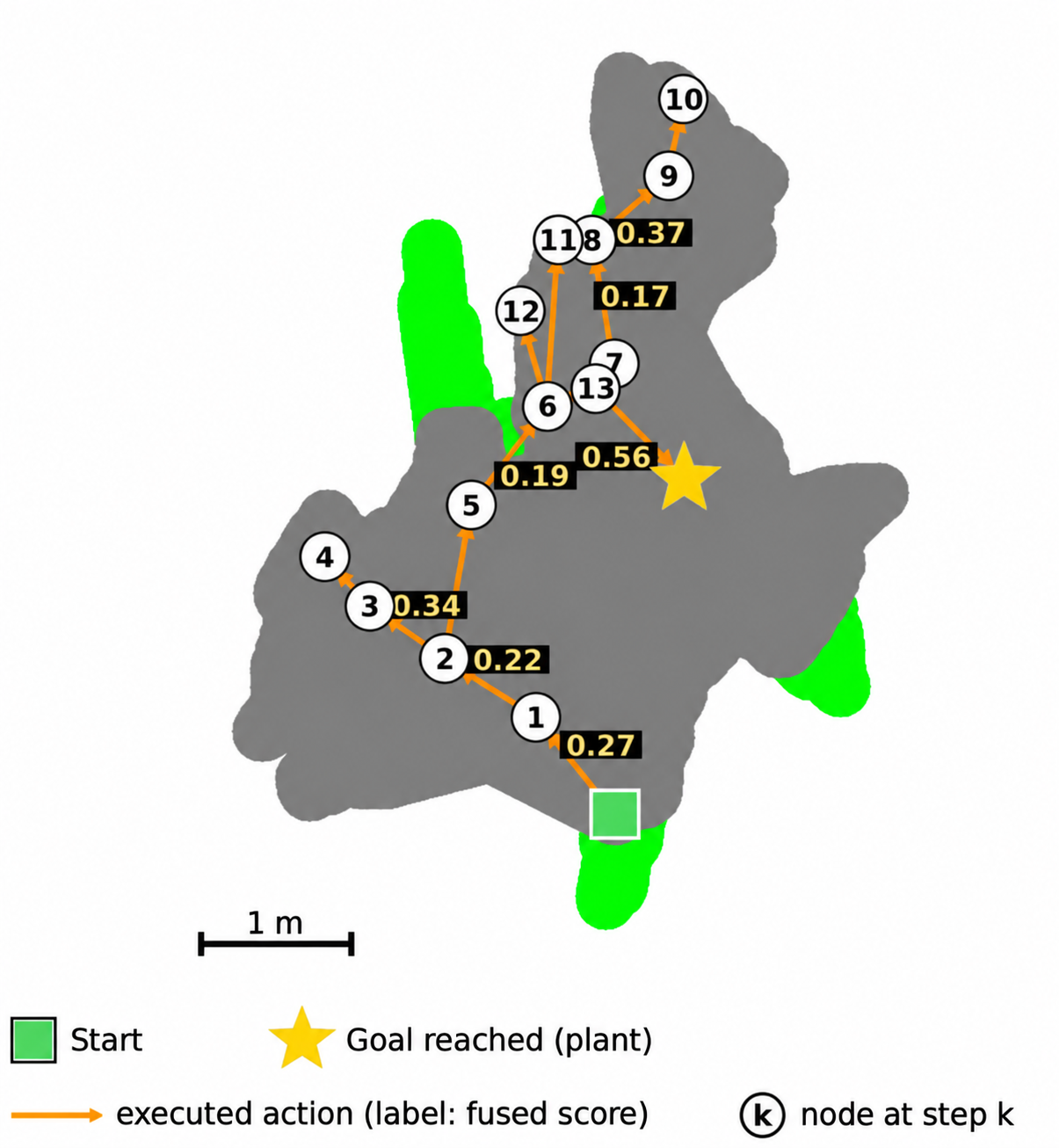}%
            }%
        }
        \caption{\cpnav: success in 18 steps.}
        \label{fig:hm3d_topdown_cpnav}
    \end{subfigure}

    \par\smallskip
    {\footnotesize
    \definecolor{hmstart}{RGB}{80,220,100}%
    \definecolor{hmgoal}{RGB}{255,215,0}%
    \definecolor{hmaction}{RGB}{255,140,0}%
    \begin{tabular}{@{}c@{}}
        \raisebox{0pt}{\fcolorbox{black}{hmstart}{\rule{0pt}{1.8mm}\rule{1.8mm}{0pt}}}~Start
        \hspace{1.5em}\textcolor{red}{$\bigstar$}~Stopped (failed)
        \hspace{1.5em}\textcolor{hmgoal}{$\bigstar$}~Goal reached (plant) \\[0.5ex]
        \textcolor{hmaction}{$\longrightarrow$}~Executed action (label: fused score)
        \hspace{1.5em}\textcircled{\scriptsize $k$}~Node at step $k$
    \end{tabular}\par
    }
    \caption{Top-down exploration maps comparing the No Calibration baseline and \cpnav on an HM3D plant-search episode. The baseline repeatedly revisits explored regions and terminates unsuccessfully after 44 steps, whereas \cpnav reaches the target plant in 18 steps with fewer revisits and less redundant exploration.}
    \label{fig:hm3d_topdown}
\end{figure*}

\subsection{VLM Token Consumption}
\label{app:tokens}

We report the average per-episode input, output, and total VLM token consumption over the 200 MP3D evaluation episodes.

\begin{table}[H]
\centering
\footnotesize
\setlength{\tabcolsep}{3pt}
\caption{Average per-episode VLM token usage.}
\label{tab:app_tokens}
\begin{tabular}{@{}lccc@{}}
\toprule
\textbf{Policy} & \textbf{Input (K)} & \textbf{Output (K)} & \textbf{Total (K)} \\
\midrule
No Calibration & 91.6  & 139.8  & 231.4 \\
Simple Set      & 82.1  & 125.4  & 207.5 \\
Prompt Set      & 81.4  & 124.3  & 205.7 \\
Ensemble Set    & 678.4 & 1125.0 & 1803.3 \\
STRIVE          & \textbf{61.3} & \textbf{38.0}   & \textbf{99.3} \\
WMNav           & 91.1  & 91.0   & 182.1 \\
\midrule
\textbf{\cpnav} & {79.8} & {121.8} & {201.6} \\
\bottomrule
\end{tabular}
\end{table}

\cpnav consumes 12.9\% fewer tokens than No Calibration, while Ensemble Set uses approximately 8.9 times more. WMNav and \cpnav both rely heavily on VLM-based reasoning, resulting in comparable token usage. STRIVE uses only 49.3\% as many tokens as \cpnav but requires heavier local computation. In our implementation, two concurrent STRIVE instances ran unstably on an RTX 4090, whereas \cpnav supported up to 40 concurrent simulation episodes and ran on the Hiwonder platform without a separate compute server.

\subsection{Qualitative Analysis of HM3D Navigation}
\label{app:hm3d_qualitative}

Figure~\ref{fig:hm3d_topdown} provides a spatial view of the exploration behavior underlying the two outcomes. The baseline trajectory contains a dense cluster of overlapping transitions near the initial region, followed by exploration of several branches. These revisits consume navigation steps without a corresponding expansion of the searched area, limiting the progress made toward the target. The \cpnav trajectory contains fewer overlapping transitions and repeated visits to explored regions, concentrating exploration on the route to the target.

This comparison highlights the importance of deciding which geometrically feasible actions warrant further exploration. The No Calibration baseline relies on the highest raw VLM score, whereas \cpnav restricts action selection to the calibrated prediction set and initiates backtracking when no candidate satisfies the threshold. This mechanism is intended to limit continued exploration along low-confidence branches while retaining alternative routes in the exploration tree. The observed reduction in redundant search is consistent with this design and provides a qualitative interpretation of the improvements in success rate and path efficiency reported in Table~\ref{tab:sim_results}. In particular, the example illustrates how the allocation of a finite action budget can influence both search efficiency and the eventual navigation outcome.

\end{document}